\documentclass[10pt,twocolumn,letterpaper]{article}

\usepackage{cvpr}
\usepackage{times}
\usepackage{epsfig}
\usepackage{graphicx}
\usepackage{amsmath}
\usepackage{amssymb}
\usepackage{makecell}

\usepackage[pagebackref=true,breaklinks=true,letterpaper=true,colorlinks,bookmarks=false]{hyperref}

\cvprfinalcopy 

\def\cvprPaperID{****} 

\ifcvprfinal\fi
\begin{document}

\title{Printing and Scanning Attack for Image Counter Forensics
}

\author{Hailey James\\
Lendbuzz\\
{\tt\small hailey.james.sorenson@lendbuzz.com}
\and
Otkrist Gupta\\
Lenduzz\\
{\tt\small otkrist.gupta@lendbuzz.com}
\and
Dan Raviv\\
Lenduzz\\
{\tt\small dan.raviv@lendbuzz.com}
}

\maketitle


\begin{abstract}
   Examining the authenticity of images has become increasingly important as manipulation tools become more accessible and advanced. Recent work has shown that while CNN-based image manipulation detectors can successfully identify manipulations, they are also vulnerable to adversarial attacks, ranging from simple double JPEG compression to advanced pixel-based perturbation. In this paper we explore another method of highly plausible attack: printing and scanning. We demonstrate the vulnerability of two state-of-the-art models to this type of attack. We also propose a new machine learning model that performs comparably to these state-of-the-art models when trained and validated on printed and scanned images. Of the three models, our proposed model outperforms the others when trained and validated on images from a single printer. To facilitate this exploration, we create a dataset of over 6,000 printed and scanned image blocks. Further analysis suggests that variation between images produced from different printers is significant, large enough that good validation accuracy on images from one printer does not imply similar validation accuracy on identical images from a different printer.
\end{abstract}

\section{Introduction}
Determining the authenticity of an image is becoming increasingly important for legal proceedings, criminal investigations, and verifying identity-supporting documents. In recent years, Convolutional Neural Networks (CNNs) have been employed to detect image manipulations, ranging from identifying splicing  and copy-move forgeries \cite{splicecopymove} to manipulations such as contrast enhancement \cite{contrast_forgery, contrastenhance}, resampling \cite{resampling_forgery}, JPEG compression \cite{jpegforgery}, Gaussian blurring \cite{gb1, gb2}, median filtering \cite{mf}, and additive white Gaussian noise \cite{BayarStamm}. Of this latter group, some may be either innocuously applied or maliciously included \cite{disguise1}.\footnote{Global manipulations such as these can be used on their own to obscure features of an image or used in tandem with another type of manipulation such as copy-move or splicing. We adopt the approach used other researchers in this area in investigating these types of global image manipulations alone (See Section \ref{section:relatedwork}).}


More recently, research in image forensics has included the presence of an adversary, a situation in which the vulnerability of CNNs has been well-studied \cite{adversarialcnn}. In regards to image forensics, consideration of these adversarial attacks have been primarily limited to pixel-based adversarial examples and JPEG double compression \cite{jpegrobust,robustmfjpeg}. In pixel-based attacks, an adversary with knowledge of the CNN model in deployment can craft an "attacked" image which appears visually identical to the original image, but is mislabeled by the CNN \cite{goodfellow2014explaining}. This problem is well known in computer vision and has been at the forefront of recent work in field. 
  However, this type of attack demands a certain level of expertise by the adversary, and is  unlikely to be employed in a majority of cases in image forensics. Even for skilled adversaries, constructing pixel-based adversarial attacks is often labor-intensive, and recent work has cast doubt on the transferability of adversarial attacks in image forensics applications \cite{analysisadvforensics}. While pixel-based adversarial attacks require at least some knowledge of the model, a low-level adversarial manipulation such as double JPEG compression requires no such knowledge \cite{jpegrobust}. In this type of attack, the images are simply JPEG compressed after the manipulation has been applied, hampering the model's ability to correctly identify post-processing methods such as additive white Gaussian noise or median filtering \cite{jpegrobust}. For this reason, building models robust to low-level, simple adversarial manipulations such as JPEG double compression, to which several manipulation detection models have been found to be vulnerable \cite{jpegrobust,robustmfjpeg,jpegrobust}, is particularly important. The goal of this paper is to investigate the vulnerability of state of the art models to another kind of low-level adversarial manipulation: printing and scanning. To our knowledge this is the first investigation into adversarial attack in digital image manipulations through printing and scanning.

\begin{figure*}[h]
  \label{fig:examples_big}
\caption{Pristine image before (left) and after (right) printing and scanning on Xerox1 (Xerox Altalink C8070 Multifunction Printer). We note that there is significant variation between the two images, similar to that introduced by the global manipulation methods with which we experimented. 
 }
\centering
\includegraphics[width=\textwidth, scale=.18]{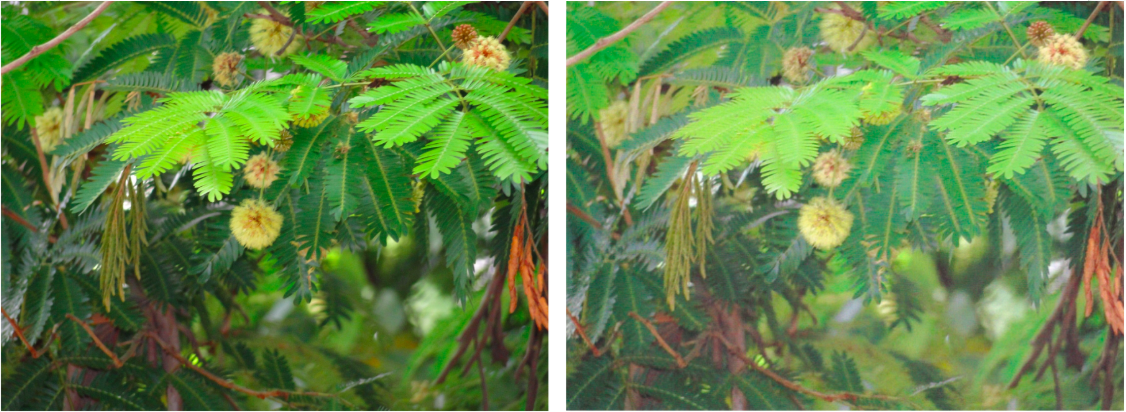}
\end{figure*}

In physical forgery, repeated printing and scanning can be used to obscure manipulations or watermarks. A document may be modified, usually non-digitally, and then repeatedly printed and scanned to disguise the manipulation artifacts. While scanning a printed document is not always related to forgery, it is reasonable to expect that state-of-the-art models be impervious to this type of post-processing, as is noted in related work in double JPEG compression \cite{jpegrobust,jpegforgery}. In addition, unlike complex pixel-based adversarial attacks, simply printing and scanning an image is both low-cost and requires little expertise, similar to JPEG compression.

In this paper, we limit our investigation to globally-applied manipulations, such as Gaussian blurring, additive white Gaussian noise, and median filtering, rather than local manipulations such as copy-move or splicing, as in related work \cite{BayarStamm}. We construct printed and scanned datasets from three different printers and experiment with two state-of-the-art models, as well as our own novel model. Related to our work is research involved in identifying camera models  \cite{cameraid} -- we additionally report results for identifying printer model. Our main contributions include the following: 
\begin{itemize}
\item We conduct the first analysis into the vulnerability of image manipulation detectors to printing and scanning, demonstrating that at least two state-of-the-art models are vulnerable to this type of highly plausible and inexpensive attack
\item We propose a novel model architecture which performs comparably than the state-of-the-art models when trained and evaluated on printed and scanned images, including performing 5\% better when trained on images from a single printer
\item We conduct an in-depth analysis on the relationship between CNN-based image manipulation detectors, including training on composite datasets, and plan to share our dataset of over 6,000 printed and scanned images with the community to facilitate further investigation
\end{itemize}




The rest of the paper is organized as follows. In Section \ref{section:relatedwork}, we give context and background through related work. In Section \ref{section:models}, we describe our novel model architecture, as well as those of the models we used for comparison. In Section \ref{section:datasets}, we describe the datasets used for training and validation. In Section \ref{section:experiments}, we explain the experiments conducted, and in Section \ref{section:results}, we discuss the results of these experiments. The paper ends in Section \ref{section:summary}, where we summarize our conclusions and suggest areas of future research.

\section{Related Work}
\label{section:relatedwork}

\begin{figure*}
\caption{Examples of manipulations before and after printing and scanning. The six manipulations refer to additive white Gaussian noise (AWGN), Gaussian blurring (GB), JPEG compression (JPEG), median filtering (MF), pristine or no manipulation (PR) and bilinear resampling (RS). We note that due to the algorithms employed, JPEG compression and resampling might be reasonably similar to the printing and scanning process. For this reason, we additionally train and evaluate the models on a restricted set of four classes only, excluding JPEG and bilinear resampling. See Table \ref{tab:Manipulation Details} for details on the parameters used for each manipulation.
}
\centering
\includegraphics[width=\textwidth, scale=.6]{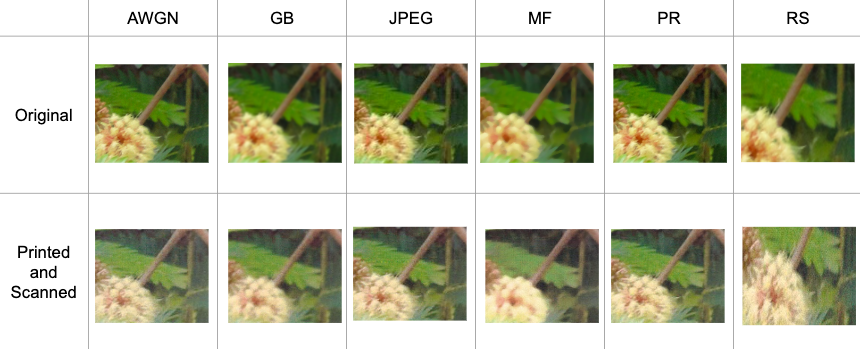}
\label{fig:examples_small}
\end{figure*}

As this paper primarily investigates manipulation detectors based on convolutional neural networks (CNNs), we provide background on CNN-based manipulation detectors. Similarly, we provide context on adversarial attacks on CNNs generally as well as specifically on CNN-based image manipulation detectors. 

Related to this work is work on detecting manipulations through inconsistencies in lighting \cite{lightning_inconsistencies} and despite various compression qualities \cite{compression}. Additionally, \cite{lightning_inconsistencies} contributes significantly to this problem area, though without examining models that leverage deep learning. \cite{compression} explores a similar problem, but without addressing specifically the problem of printing and scanning in relation to CNN-based detectors, and is thus complementary to this work.

\subsection{Deep Learning for Image Forensics}

 Recent methods in image forensics techniques leveraging deep learning have reached impressive performance. In 2015, a CNN-based classifier was proposed for detecting median filtering in images \cite{mfcnn}. Building on this work, \cite{BayarStamm} proposed CNN-based model with the addition of a "constrained convolutional layer", or a layer constrained to learn the high-pass features of an image by attempting to predict a central pixel based on its neighbors. This serves to suppress the image content while learning the manipulation fingerprint, drawing inspiration from steganalysis rich model (SRM) filters in steganalysis \cite{FridrichSRM}.  
 
 More recently, it has been shown that third order subtractive pixel analysis matrix (S3SPAM) features can be learned by a simple shallow CNN, and can employ transfer learning to achieve good performance on little training data \cite{Cozzolino_2017}. In addition to directly detecting manipulations, a deep learning method for analyzing the image processing history as an important component for image forensics has been proposed, as the processing history pipeline can affect the accuracy of other forensic tools \cite{FridrichHistory}.
 
The performance achieved by constrained convolutional layers and particularly deep networks is particularly impressive. These techniques serve as inspiration for our proposed model, and we thus compare our proposed model with models that leverage these modifications.
 
 \begin{table*}[t]
\caption{\label{tab:Dataset Descriptions}Descriptions and sizes of each dataset used for training and validation. Size refers to the number of 299x299 or 256x256 image blocks in each dataset, which is then split in 75\% training and 25\% validation. X1 and X2 refer to Xerox1 and Xerox2 printers (Section \ref{section:printscan}), respectively. The labels refer to the labels used when training and evaluating on each dataset}
\begin{tabular}{c|l|l|l}
Dataset Name                   & Description                                                                            & Size    & Labels                        \\ \hline
Original                  & IFSTC dataset after six manipulations                                                  & 198,624 & \makecell[l]{awgn, gb, mf, \\ pr (jpeg), (rs)} \\
Xerox1                & \makecell[l]{Images from IFSTC dataset with \\ manipulations after being printed \\ and scanned on Xerox1}  & 2,142   & \makecell[l]{awgn, gb, mf, \\ pr (jpeg), (rs)} \\
Composite Printers     & Combined set of images from each printer (balanced)                                   & 6,426   & \makecell[l]{awgn, gb, mf, \\ pr (jpeg), (rs)} \\
Composite Full         & \makecell[l]{Combined set of images from each\\ printer plus original IFSTC images (balanced)}        & 8,568   & \makecell[l]{awgn, gb, mf, \\ pr (jpeg), (rs)} \\
Printer Identification & Pristine images after being printed and scanned by all three printers                  & 3,213   & Dell, X1, X2                  \\
JPEG Compression       & IFSTC dataset with JPEG compression (QF=80) on all images                              & 198,624 & \makecell[l]{awgn, gb, mf, \\ pr (jpeg), (rs)}
\end{tabular}
\end{table*}

\subsection{Adversarial Attacks on CNNs}
The vulnerability of CNNs to adversarial attacks has been well documented \cite{goodfellow2014explaining, papernot2016limitations}. Adversarial noise can be designed in such a way that, when added to the image, can retain visual quality while misleading the classifier. For example, Fast Gradient Sign Method (FGSM) \cite{goodfellow2014explaining} leverages the differentiability of the loss function, assumed to be known to the adversary. The method proposes altering each pixel based on the gradient of the loss with respect to the original pixels in the input image. These changes small are enough such that the resulting image is visually nearly identical to the original, but are large enough  cumulatively to increase the loss such as to impair the classification. Similarly, projected gradient descent (PGD) \cite{pgd} seeks a perturbation that maximises the loss on a specific input while keeping the perturbation size smaller than a given epsilon. DeepFool \cite{moosavi2016deepfool} uses a local linearization of the classifier to approximate the decision boundary and alter the images accordingly. The Jacobian-based Saliency Map Attack (JSMA) \cite{papernot2016limitations} uses a greedy iterative procedure, altering only the pixels which contribute most to the correct classification as identified by a saliency map. Each of these pixel-based adversarial attacks, while effective, requires at least partial knowledge of the network used for image manipulation detection. In contrast, low-level adversarial attacks such as JPEG compression or printing and scanning, the subject of this paper, require no such knowledge. 
 
\subsection{Adversarial Attacks in Image Forensics}

While CNN-based classifiers have achieved high performance on benchmark image forensic tasks, recent research in computer vision has demonstrated that CNN-based manipulation detectors, like CNNs more broadly, are highly vulnerable to adversarial attacks. For example, in \cite{AdvGAN}, the authors demonstrate that a GAN-based architecture can conceal 3x3 median filtering manipulation, one of the manipulations we explore in this paper. This type of adversarial attack causes a detector to label the image as non-manipulated, including for the CNN-based detectors proposed in \cite{BayarStamm} and \cite{cnnmf}. Additionally, a method of adversarial attack based on small pixel-based distortions has been proposed for fooling global image manipulation detectors \cite{pixeldomainadv}. However, \cite{transferadv} notes that unlike in most pattern recognition tasks, pixel-based adversarial attacks such as Fast Gradient Sign Method (FGSM) \cite{goodfellow} and Jacobian-based Saliency Map Attack
(JSMA) \cite{Papernot}, are not for the most part transferable between manipulation detection models. 

Recent work has explored the vulnerability of image manipulation detectors to low-resolution median filtering \cite{mfnet} and  JPEG compression \cite{jpegrobust, robustmfjpeg, multipurposecnn}. To our knowledge, ours is the first paper to examine model vulnerability to printing and scanning.  

\section{Models}
\label{section:models}

Here we describe our proposed model architecture for improved performance on printed and scanned images. We compare our model's performance with the model proposed in \cite{BayarStamm}, the inspiration for the constrained convolutional layer. We additionally compare our model with XceptionNet (Xception) \cite{xception}, as it and our proposed model have nearly identical number of parameters and similar architecture, so the difference in performance cannot be attributed to increased network capacity. 
\begin{figure*}[h]
\caption{Proposed network architecture. The first layer is a constrained convolutional layer to extract SRM features, followed by an deep architecture with separable convolutional layers to improve generalization. The last layer is either a 4x1 vector (for four classes) or a 6x1 vector (for all six classes).}
\centering
\includegraphics[width=\textwidth, scale=.18]{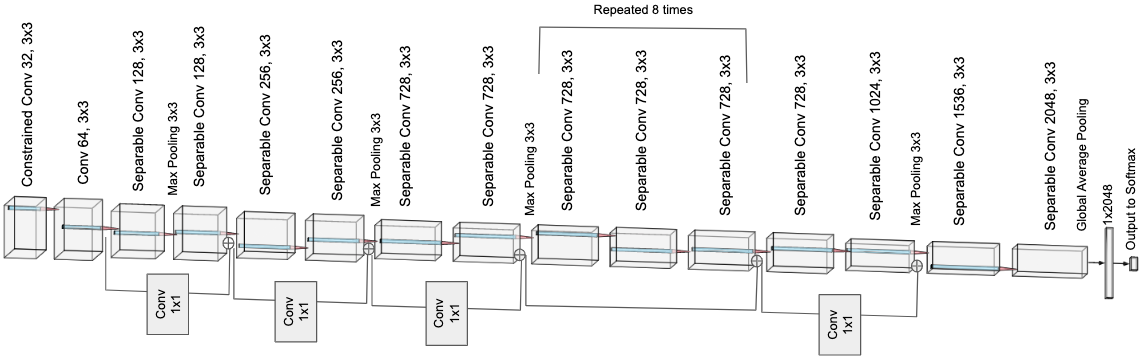}
  \label{fig:network_diagram}
\end{figure*}

\subsection{Proposed Model}

Our proposed architecture consists of one constrained convolutional layer \cite{BayarStamm}, 1 convolutional layer, 34 separable convolutional layers, 5 pooling layers (4 max pooling, 1 global average pooling), and a final fully connected layer (see Figure \ref{fig:network_diagram}). Each convolutional layer was followed by ReLU activation, and max pooling layers were performed with a stride of 2x2.

In the constrained convolutional layer, a 5x5 filter is employed in which the sum of all the weights is constrained to be zero \cite{BayarStamm}. Specifically, the center pixel is predicted by the rest of the pixels in the field, and the output of the filter can be interpreted as the prediction error, as suggested by research in steganalysis \cite{FridrichHistory}.  Specifically, the weights in the filter are constrained such that: 

\[ \begin{cases} 
     w(0,0) & =-1\\
      \sum_{l,m \neq 0 }w(l,m) & =1 \\
   \end{cases}
\]
where $w$ refers to the weight, and $l$ and $m$ refer to the coordinates in the filter, where $0,0$ is the central weight. 

The purpose of the constrained convolutional layer is to constrain the model to learn image manipulation fingerprints, rather than image content and higher order features, such as those useful for object recognition and classification tasks. The prediction error fields are then used as low-level forensic trace features by the rest of the network to assist in classifying global image manipulation detection. 

For the separable convolutional layers, a spatial convolution is performed independently for each channel and is followed by a point-wise or 1x1 convolution, as proposed in \cite{xception}. These components decrease the number of free parameters allowing the deep network to learn effectively even with a small training set, making it particularly appropriate for our investigation. 

In this approach, we hope to leverage both the SRM-like features produced by the convolutional layer as well as the improved generalization ability provided by the added depth and separable layers.

\subsection{Bayar2016}

Proposed in 2016, the constrained convolution method of image manipulation detection, hereafter referred to at Bayar2016, proposes a three-layer CNN, with two max-pooling layers and three fully-connected layers (including the initial constrained convolutional layer) \cite{BayarStamm}. This model demonstrates impressive results in discerning between the six manipulations investigated in this paper using the dataset described in the next section, achieving 99.9\% validation accuracy. 

\subsection{Xception}

In addition to a the Bayar2016 shallow network, recent work has demonstrated that increasing network depth can dramatically improve model generalization. To compare with a model of similar depth that also uses separable convolutional layers, we experiment with XceptionNet, a deep network comprising of 42 layers, including separable convolutional layers \cite{xception}. The network design is built upon Inception architecture \cite{inception}, with the innovation of separable filters. Similar to Bayar2016, this model also achieves near 99\% accuracy on the dataset described in \cite{BayarStamm} before printing and scanning.


\section{Datasets}
\label{section:datasets}

\begin{table}[t]
\begin{tabular}{l|l}
Manipulation                         & Hyperparameters                                                                           \\ \hline
\makecell[l]{Additive White Gaussian Noise \\ (AWGN)} & $\sigma$ = 2.0                                                               \\
Gaussian Blurring (GB)               & \begin{tabular}[c]{@{}l@{}}kernal size = (5,5)\\ $\sigma$ = 1.1\end{tabular} \\
JPEG compression (JPEG)              & quality = 70                                                                              \\
Median Filtering (MF)                & kernal size = (5,5)                                                                       \\
Pristine (PR)                        & none                                                                                      \\
Bilinear Resampling (RS)             & ratio = 1.5                                                                  \\            
\end{tabular}
\caption{\label{tab:Manipulation Details}Parameter specifications for each manipulation type. We used the same parameters as in \cite{BayarStamm} for fair comparison. See Section \ref{section:manipulations} for details on the manipulations.}
\end{table}

For accurate comparison, we follow the procedure described in \cite{BayarStamm}, using images from the first IEEE IFSTC Image Forensics Challenge \cite{ifstc}. The portion of the dataset used consists of 3,334 images of size 1024x768, which was further split into training, validation and testing data. The images are captured from several different digital cameras of both indoor and outdoor scenes.

\subsection{Printing and Scanning}
\label{section:printscan}
We used three different printers and one scanner to create a dataset of printed and scanned images: one Dell S3845CDN Laser Multifunction Printer, one Xerox Altalink C8070 Multifunction Printer, and one Xerox WorkCentre 7970 Multifunction Printer, which we refer to as Dell, Xerox1 and Xerox2 respectively hereafter. We printed 50 images of each manipulation type on each printer and used the Dell scanner to scan each image (see Figure \ref{fig:examples_big}). After scanning and extracting the images from the resulting pdfs, the image sizes were 1700 x 2200 pixels, which was then center-cropped to 1536 x 1792 to remove the white border added by the scanning process. Each image was then split into 42 299x299 blocks (or 256x256 blocks for Bayar2016), resulting in 2,142 image blocks of each class from each printer (see Figure \ref{fig:examples_small}). We limited our data creation to 900 full-page color images both for budget constraints and environmental concerns; creating a synthetic dataset through printing and scanning simulation may be an avenue of future work.

\subsection{Manipulations}
\label{section:manipulations}

Again following the procedure described in  \cite{BayarStamm}, we manipulated each image with each of six manipulation types: additive white Gaussian noise (AWGN), Gaussian blurring (GB), JPEG compression (JPEG), median filtering (MF), re-sampling (RS) and retaining the pristine image (PR).

\begin{itemize}
\item Additive white Gaussian noise constructs a noise matrix of the same shape as the image according to a normal distribution with a given sigma value and adds this matrix to the original image. The result is then normalized to values between 0 and 255. 
\end{itemize}
\begin{itemize}
\item Gaussian blurring blurs the image using a Gaussian filter by convolving the input image using a given kernal.
\item JPEG compression is a lossy compression method which compresses the image through converting the color map, down-sampling and Discrete Cosine Transform (DCT).
\item Median filtering replaces each pixel with the median value of the neighboring pixels using a given kernal area.
\item Bilinear resampling works similarly, resizing the image using the distance-weighted average of the neighboring pixels to estimate the new pixel value. 
\end{itemize}
See Table \ref{tab:Manipulation Details} for manipulation parameter details.

\section{Experiments}
\label{section:experiments}

We trained each model (our proposed model, Bayar2016, and Xception) on a variety of training sets and evaluated each trained model on multiple validation datasets (See Table  \ref{tab:Dataset Descriptions}).

We first investigated the extent to which our selected models can correctly classify the validation images after printing and scanning. We trained each model on the  original dataset (before printing and scanning) with all six classes: additive white gaussian noise (AWGN), gaussian blurring (GB), JPEG compression (JPEG), median filtering (MF), bilinear resampling (RS) and pristine or no manipulation (PR). For a more complete analysis, we removed the bilinear resampling (RS) and JPEG compression (JPEG) classes from the training and validations sets and retrained the models, as these two classes could intuitively be considered similar to changes introduced during the printing and scanning process (see Table \ref{tab:orig_results}).

Second, we explored countering this vulnerability by training on the printed and scanned image blocks \cite{goodfellow}. We trained each model on the printed and scanned image blocks from a single printer. The dataset (see Table \ref{tab:Dataset Descriptions}, Xerox1) consists of 50 full images (1700 x 2200 pixels), which were then divided into 299x299 for our proposed model and Xception, and 256x256 for the Bayar2016 model. This resulted in 2,142 image blocks for each dataset, which was divided into training and validation sets of size 1722 and 420 respectively, using only the central images to avoid including border artifacts from the scanning process. 

Third, we created composite datasets, one consisting of all printed and scanned image blocks (from all three printers), and the other consisting of all printed and scanned image blocks as well as a number of image blocks from the original dataset (before printing and scanning), at a size equivalent to those from one of the three printers. The first composite dataset, which we refer to as Composite Printers, consists of 6,426 image blocks (printed and scanned only), while the second consists of 8,568 image blocks (75\% printed and scanned, 25\% original). The goal of this experiment was to evaluate if the poor accuracy fitting the printed and scanned data could be mitigated by dramatically increasing the size of the training set.

Finally, we evaluated the performance of each of the models on identifying the printer of printed and scanned images (see Table \ref{tab:Dataset Descriptions}, Printer Identification). 

\subsection{Hyperparameters}

For Bayar2016, we used a batch size of 64, an initial learning rate of 0.01, Stochastic Gradient Descent (SGD) with momentum 0.95, weight decay 0.0005, gamma 0.7, and step size 6. 

We used similar hyperparameters for Xception and our proposed model. Specifically, for both models, we use the pre-trained weights from the network as trained on ImageNet. We again used SGD, and inferred the batch size and learning rate at training time based on the number of GPUs, using  $$batch\_size = 4  \times num\_gpus$$ for the batch size and 0.01 for the initial learning rate. We use momentum 0.9 and weight decay 0.0005. For learning rate decay, we use polynomial decay as described in \cite{psp}. For each model, we trained until the validation accuracy plateaued or began to fall. 

Following the original methodology for Bayar2016, we retain only the green color layer of each image and divide into 256x256 non-overlapping blocks, retaining nine central blocks. For our proposed model and for Xception, we retain all three color channels and split the images into 299x299 non-overlapping blocks, according to the input size of the original architecture. 

\section{Results and Discussion}
\label{section:results}

\subsection{Print-Scan Manipulation}

To evaluate the general vulnerability of each of the models to printed and scanned images, we trained on the original IFSTC dataset (before printing and scanning) and evaluated each model on validation sets from each of the three printers. When we evaluated the models on the printed and scanned validation sets, we found that each model performed only slightly better than random. 

We additionally removed the bilinear resampling and JPEG compression classes, and found that the resulting models are similarly unable to correctly classify the remaining four manipulations, still performing at or below random. We additionally note that the models perform worse on the printed and scanned validation images than on the validation images after JPEG compression, a known vulnerability of these types of models, indicating that printing and scanning may be more effective at masking the manipulations \cite{compression} (see Table \ref{tab:orig_results}). 


\begin{table}[]
\caption{\label{tab:orig_results} Validation accuracy for various validation sets after training on IFSTC dataset. We note that although all three models perform exceptionally well on the original IFSTC dataset, each performs little better than random when evaluated on images from any of the three printers. Because JPEG compression and bilinear resampling could be reasonably inferred to be similar to printing and scanning, we remove these classes and train and evaluate on a restricted set of four classes (4c) (See Section \ref{section:experiments}). Despite this restricted set of manipulations, however, the models perform no better than random.}
\begin{tabular}{l|c|c|c}
         & \multicolumn{1}{l|}{Bayar2016} & \multicolumn{1}{l|}{Xception}   & \multicolumn{1}{l}{ \makecell{Proposed \\ Model}} \\ \hline
Original (6c) & \multicolumn{1}{c|}{0.9979}   & \multicolumn{1}{c|}{0.9916} & 0.993                                      \\
Dell (6c)     & \multicolumn{1}{c|}{0.1643}   & \multicolumn{1}{c|}{0.1632} & 0.1673                                      \\
Xerox1 (6c)   & \multicolumn{1}{c|}{0.1976}   & \multicolumn{1}{c|}{0.201} & 0.1827                                      \\
Xerox2 (6c)   & \multicolumn{1}{c|}{0.1972}   & \multicolumn{1}{c|}{0.202} & 0.1953                                      \\ \hline

Original (4c) & \multicolumn{1}{c|}{0.9948}   & \multicolumn{1}{c|}{0.9954}    & 0.997                                      \\
Dell (4c)     & \multicolumn{1}{c|}{0.2571}   & \multicolumn{1}{c|}{0.223}     &  0.2347                                      \\
Xerox1 (4c)   & \multicolumn{1}{c|}{0.2411}   & \multicolumn{1}{c|}{0.246}     & 0.2367                                      \\
Xerox2 (4c)   & \multicolumn{1}{c|}{0.2387}   & \multicolumn{1}{c|}{0.255}     & 0.2393                                      \\
JPEG (4c)     & \multicolumn{1}{c|}{0.4255}   & \multicolumn{1}{c|}{0.5126}    & 0.4825                                     
\end{tabular}
\end{table}

\begin{table}[]
\caption{\label{tab:hawk_results}Validation accuracy for various validation sets after training on Xerox1 dataset (see Table \ref{tab:Dataset Descriptions}, Xerox1). We trained each model on images from only the Xerox1 dataset, or images after being printed and scanned on the first Xerox printer. We find that while no model is able to perfectly fit the printed and scanned dataset, our proposed models significantly outperforms the current state-of-the-art models.  We also note that transferability to other printers remains weak, indicating significant variance between the printers. Here 4c indicates that we used the restricted set of manipulations (AWGN, GB, MF, and PR) (See Section \ref{section:experiments}).}
\begin{tabular}{l|c|c|c}
         & \multicolumn{1}{l|}{Bayar2016} & \multicolumn{1}{l|}{Xception} & \multicolumn{1}{l}{\makecell{Proposed \\ Model}} \\ \hline
Xerox1 (4c)   & 0.7036                         & 0.666                         & \textbf{0.753}                           \\
Dell (4c)     & 0.3018                         & 0.482                         & 0.456                                    \\
Original (4c) & 0.2342                         & 0.3873                        & 0.3649                                   \\
Xeros2 (4c)   & 0.4738                         & 0.611                         & 0.572                                    \\
JPEG (4c)     & 0.2418                         & 0.3848                        & 0.364                                   
\end{tabular}
\end{table}

\begin{table}[]
\caption{\label{tab:composite_printers_results} Validation accuracy for various validation sets after training on the Composite Printers dataset. One possible explanation for the poor validation accuracy on a single printer could be the small size of the dataset. To investigate this, we combine the images from all three printers for training, but note that performance on a single printer does not improve. Here 4c indicates that we used the restricted set of manipulations (AWGN, GB, MF, and PR) (See Section \ref{section:experiments}).}
\begin{tabular}{l|c|c|c}
         & \multicolumn{1}{l|}{Bayar2016} & \multicolumn{1}{l|}{Xception} & \multicolumn{1}{l}{\makecell{Proposed \\ Model}} \\ \hline

Dell (4c)     & 0.6506                         & 0.649                         & \textbf{0.713 }                                 \\
Xerox1 (4c)   &\textbf{ 0.7001}                         & 0.626                         & \textbf{0.696 }                                     \\
Xeros2 (4c)   & 0.5381                         & 0.623                         & \textbf{0.663}                                      \\
JPEG (4c)     & 0.2643                         & 0.2902                        & 0.2601
\\
Original (4c) & 0.2617                         & 0.2847                        & 0.2449
\end{tabular}

\end{table}


\begin{table}[]
\caption{\label{tab:composite_full_results} Validation accuracy for various validation sets after training on the Composite Printers dataset. For a complete analysis, we add additional image blocks (blocks before printing and scanning) to the composite dataset, but again find that performance does not improve. Here 4c indicates that we used the restricted set of manipulations (AWGN, GB, MF, and PR) (See Section \ref{section:experiments}).}
\begin{tabular}{l|c|c|c}
         & \multicolumn{1}{l|}{Bayar2016} & \multicolumn{1}{l|}{Xception} & \multicolumn{1}{l}{\makecell{Proposed \\ Model }} \\ \hline

Dell (4c)     & 0.6339                         & \textbf{0.662}                           & \textbf{0.661}                                      \\
Xerox1 (4c)   & 0.6982                         & 0.632                           & 0.674                                      \\
Xerox2 (4c)   & 0.5637                         & 0.602                           & 0.696                                      \\
JPEG (4c)     & 0.519                          & 0.6374                           & 0.4972   
\\
Original (4c) & 0.8063                         & 0.0.9259                           & 0.9629                                                                   
\end{tabular}
{}
\end{table}


\begin{table}[]
\caption{\label{tab:printer_id} Validation accuracy for printer identification by model. We investigate the variation of the images between printers by training each model to discern between printers. The high accuracy indicates that the images produces by each printer vary significantly.}
\begin{tabular}{l|lll}
                  & \multicolumn{1}{l|}{Bayar2016}        & \multicolumn{1}{l|}{Xception}                    & \multicolumn{1}{l}{\makecell{Proposed \\ Model}}     \\ \hline
\makecell[l]{Printer \\ Identification} & \multicolumn{1}{c|}{0.9048} & \multicolumn{1}{c|}{0.956} & \multicolumn{1}{c}{0.9533}
\end{tabular}
{}
\end{table}

\subsection{Cross-Training on Printed and Scanned Examples }

We additionally trained each model on printed and scanned images from an individual printer (Xerox1) (See Section \ref{section:printscan}). 

We note that Bayar2016 and Xception achieve accuracies 66.6\% and 70.4\% respectively, while our proposed model is able to achieve an accuracy of 75.3\%.  It also appears that training on one printer does not lend itself to similar validation accuracy on examples from another printer, even of the same make. (See Table \ref{tab:hawk_results}).

\subsection{Composite Training}

To compensate for the small size of the dataset for each printer alone, we created a composite dataset, consisting of all of the printed and scanned examples (total size 6,426 blocks), which we refer to as Composite Printers. However, we found that training on this composite dataset did not improve validation performance on any single printer compared with training on images from that printer alone. While this is possibly due to a still insufficiently large training dataset, it also likely provides further evidence that the difference between printers and scanners may be significant enough to preclude fitting a general printed and scanned dataset (see Table \ref{tab:composite_printers_results}).

For completion, we additionally created another composite dataset, which we refer to as Composite Full, which consists of the same composition as Composite Printers plus an equivalent number of examples from the original dataset (total size 8,568), and found similar results. (See Table \ref{tab:composite_full_results}). 


\subsection{Printer Identification}

For comparison with work on  camera model identification, we additionally experimented with printer identification on each of the three printers using the discussed models, and found that the models could distinguish between images from the printers with up to 95\% accuracy. This is particularly impressive considering the accuracies were achieved using a relatively small set of training data (2,410 image blocks) and without any additional metadata (see Table \ref{tab:printer_id}), indicating significant variance between the artifacts introduced by each printer \cite{cameraid}.

\section{Summary}
\label{section:summary}

We investigated the robustness of current state-of-the-art image manipulation detection models in the context of printing and scanning, and found that these models perform poorly on printed and scanned image data. We proposed our own novel model architecture, which performs ~5\% better than the state-of-the-art models when trained and evaluated on images from a single printer. We constructed a dataset of over 6,000 printed and scanned image blocks which we plan to release to the community for further investigation.

That current state-of-the-art models are vulnerable to printing and scanning is an important finding given the availability and ease of printing and scanning images versus constructing complex adversarial examples. 

Further analysis suggest that the variability between images produced by each printer is large, significant enough for the models to easily distinguish between printers and for models trained on a single printer to generalize poorly to images from another printer. This conclusion may create additional challenges in designing models robust to printing and scanning, and sets it apart from work on creating models robust to more uniform and predictable JPEG compression. Future work may include developing methods to simulate printing and scanning in order to create a larger datasets for training the models. 

\begin{figure}[]
  \label{fig:bayar_orig_6c}
\caption{Confusion matrix for Bayar2016 trained on Original IFSTC, evaluated on Xerox1 (see Table \ref{tab:Dataset Descriptions}, Xerox1). We note that despite the high reported validation accuracy on the original dataset, the model struggles to distinguish between the classes after printing and scanning.}
\centering
\includegraphics[ scale=.3]{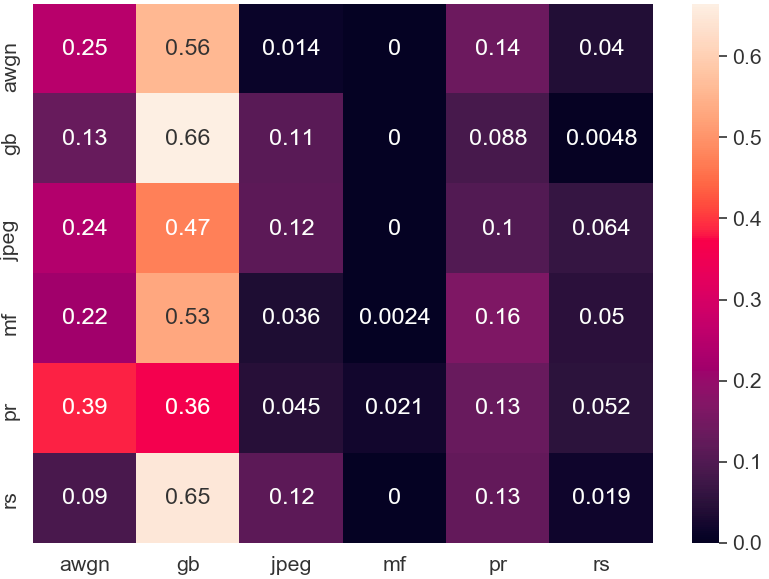}
\end{figure}

\begin{figure}[]
  \label{fig:bayar_orig_4c}
\caption{Confusion matrix for Bayar2016 trained on Original IFSTC (without RS and JPEG), evaluated on Xerox1 (See Table \ref{tab:Dataset Descriptions}). We investigate the model's performance after removing bilinear resampling and JPEG compression, but find that it still performs little better than random.}
\centering
\includegraphics[scale=.3]{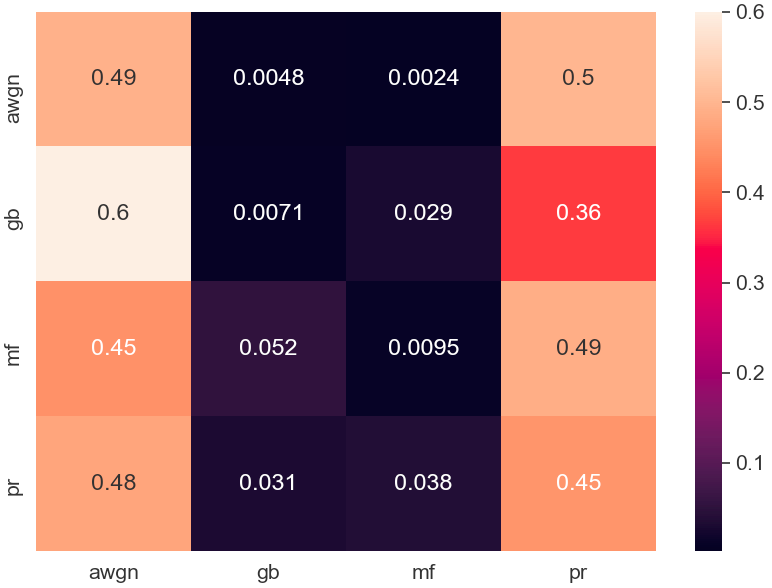}
\end{figure}

{\small
\bibliographystyle{ieee_fullname}
\bibliography{egbib}

\begin{thebibliography}{1}\itemsep=-1pt

\bibitem{Alpher02}
A. Alpher.
\newblock Frobnication.
\newblock {\em Journal of Foo}, 12(1):234--778, 2002.

\bibitem{Alpher03}
A. Alpher and J.~P.~N. Fotheringham-Smythe.
\newblock Frobnication revisited.
\newblock {\em Journal of Foo}, 13(1):234--778, 2003.

\bibitem{Alpher04}
A. Alpher, J.~P.~N. Fotheringham-Smythe, and G. Gamow.
\newblock Can a machine frobnicate?
\newblock {\em Journal of Foo}, 14(1):234--778, 2004.

\bibitem{Authors20}
Authors.
\newblock The frobnicatable foo filter, 2020.
\newblock Face and Gesture submission ID 324. Supplied as additional material
  {\tt fg324.pdf}.

\bibitem{Authors20b}
Authors.
\newblock Frobnication tutorial, 2020.
\newblock Supplied as additional material {\tt tr.pdf}.

\end{thebibliography}


\begin{thebibliography}{10}\itemsep=-1pt

\bibitem{jpegrobust}
M. Barni, A. Costanzo, E. Nowroozi, and B. Tondi.
\newblock Cnn-based detection of generic contrast adjustment with jpeg
  post-processing.
\newblock {\em 2018 25th IEEE International Conference on Image Processing
  (ICIP)}, Oct 2018.

\bibitem{BayarStamm}
Belhassen Bayar and Matthew~C. Stamm.
\newblock A deep learning approach to universal image manipulation detection
  using a new convolutional layer.
\newblock {\em ACM Workshop on Information Hiding and Multimedia Security},
  pages 5--10, 2016.

\bibitem{FridrichHistory}
Mehdi Boroumand and Jessica~J. Fridrich.
\newblock Deep learning for detecting processing history of images.
\newblock In {\em Media Watermarking, Security, and Forensics}, 2018.

\bibitem{gb2}
Gang Cao, Yao Zhao, and Rongrong Ni.
\newblock Edge-based blur metric for tamper detection.
\newblock {\em Journal of Information Hiding and Multimedia Signal Processing},
  1(1):20--27, 2010.

\bibitem{mf}
Gang Cao, Yao Zhao, Rongrong Ni, Lifang Yu, and Huawei Tian.
\newblock Forensic detection of median filtering in digital images.
\newblock In {\em 2010 IEEE International Conference on Multimedia and Expo},
  pages 89--94. IEEE, 2010.

\bibitem{disguise1}
G. {Cao}, Y. {Zhao}, R. {Ni}, L. {Yu}, and H. {Tian}.
\newblock Forensic detection of median filtering in digital images.
\newblock In {\em 2010 IEEE International Conference on Multimedia and Expo},
  pages 89--94, 2010.

\bibitem{adversarialcnn}
Nicholas Carlini and David Wagner.
\newblock Adversarial examples are not easily detected: Bypassing ten detection
  methods.
\newblock In {\em Proceedings of the 10th ACM Workshop on Artificial
  Intelligence and Security}, pages 3--14, 2017.

\bibitem{mfcnn}
Jiansheng Chen, Xiangui Kang, Ye Liu, and Z. Wang.
\newblock Median filtering forensics based on convolutional neural networks.
\newblock {\em Signal Processing Letters, IEEE}, 22:1849--1853, 11 2015.

\bibitem{cnnmf}
Jiansheng Chen, Xiangui Kang, Ye Liu, and Z. Wang.
\newblock Median filtering forensics based on convolutional neural networks.
\newblock {\em Signal Processing Letters, IEEE}, 22:1849--1853, 11 2015.

\bibitem{multipurposecnn}
Yifang Chen, Xiangui Kang, Y.Q. Shi, and Z. Wang.
\newblock A multi-purpose image forensic method using densely connected
  convolutional neural networks.
\newblock {\em Journal of Real-Time Image Processing}, 16, 03 2019.

\bibitem{xception}
Francois Chollet.
\newblock Xception: Deep learning with depthwise separable convolutions.
\newblock {\em 2017 IEEE Conference on Computer Vision and Pattern Recognition
  (CVPR)}, Jul 2017.

\bibitem{Cozzolino_2017}
Davide Cozzolino, Giovanni Poggi, and Luisa Verdoliva.
\newblock Recasting residual-based local descriptors as convolutional neural
  networks.
\newblock {\em Proceedings of the 5th ACM Workshop on Information Hiding and
  Multimedia Security - IHMMSec ’17}, 2017.

\bibitem{compression}
Mathieu Dejean-Servi{\`e}res, Karol Desnos, Kamel Abdelouahab, Wassim
  Hamidouche, Luce Morin, and Maxime Pelcat.
\newblock Study of the impact of standard image compression techniques on
  performance of image classification with a convolutional neural network.
\newblock 2017.

\bibitem{jpegforgery}
Hany Farid.
\newblock Exposing digital forgeries from jpeg ghosts.
\newblock {\em IEEE transactions on information forensics and security},
  4(1):154--160, 2009.

\bibitem{FridrichSRM}
Jessica Fridrich and Jan Kodovsky.
\newblock Rich models for steganalysis of digital images.
\newblock {\em IEEE Transactions on Information Forensics and Security},
  7:868--882, 06 2012.

\bibitem{goodfellow2014explaining}
Ian~J Goodfellow, Jonathon Shlens, and Christian Szegedy.
\newblock Explaining and harnessing adversarial examples.
\newblock {\em arXiv preprint arXiv:1412.6572}, 2014.

\bibitem{goodfellow}
Ian~J. Goodfellow, Jonathon Shlens, and Christian Szegedy.
\newblock Explaining and harnessing adversarial examples, 2014.

\bibitem{analysisadvforensics}
Diego Gragnaniello, Francesco Marra, Giovanni Poggi, and Luisa Verdoliva.
\newblock Analysis of adversarial attacks against cnn-based image forgery
  detectors.
\newblock {\em 2018 26th European Signal Processing Conference (EUSIPCO)}, Sep
  2018.

\bibitem{gb1}
Dun-Yu Hsiao and Soo-Chang Pei.
\newblock Detecting digital tampering by blur estimation.
\newblock In {\em First International Workshop on Systematic Approaches to
  Digital Forensic Engineering (SADFE'05)}, pages 264--278. IEEE, 2005.

\bibitem{lightning_inconsistencies}
M.~K. {Johnson} and H. {Farid}.
\newblock Exposing digital forgeries in complex lighting environments.
\newblock {\em IEEE Transactions on Information Forensics and Security},
  2(3):450--461, 2007.

\bibitem{AdvGAN}
Dongkyu Kim, Han-Ul Jang, Seung-Min Mun, Sunghee Choi, and Heung-Kyu Lee.
\newblock Median filtered image restoration and anti-forensics using
  adversarial networks.
\newblock {\em IEEE Signal Processing Letters}, 25:278--282, 2018.

\bibitem{pgd}
Aleksander Madry, Aleksandar Makelov, Ludwig Schmidt, Dimitris Tsipras, and
  Adrian Vladu.
\newblock Towards deep learning models resistant to adversarial attacks.
\newblock {\em arXiv preprint arXiv:1706.06083}, 2017.

\bibitem{moosavi2016deepfool}
Seyed-Mohsen Moosavi-Dezfooli, Alhussein Fawzi, and Pascal Frossard.
\newblock Deepfool: a simple and accurate method to fool deep neural networks.
\newblock In {\em Proceedings of the IEEE conference on computer vision and
  pattern recognition}, pages 2574--2582, 2016.

\bibitem{papernot2016limitations}
Nicolas Papernot, Patrick McDaniel, Somesh Jha, Matt Fredrikson, Z~Berkay
  Celik, and Ananthram Swami.
\newblock The limitations of deep learning in adversarial settings.
\newblock In {\em 2016 IEEE European symposium on security and privacy
  (EuroSP)}, pages 372--387. IEEE, 2016.

\bibitem{Papernot}
Nicolas Papernot, Patrick McDaniel, Somesh Jha, Matt Fredrikson, Z.~Berkay
  Celik, and Ananthram Swami.
\newblock The limitations of deep learning in adversarial settings.
\newblock {\em 2016 IEEE European Symposium on Security and Privacy (EuroSP)},
  Mar 2016.

\bibitem{resampling_forgery}
Alin Popescu and Hany Farid.
\newblock Exposing digital forgeries by detecting traces of re-sampling.
\newblock {\em IEEE Transactions on Signal Processing}, 53:758--767, 02 2005.

\bibitem{ifstc}
A.C. Popescu and H. Farid.
\newblock Exposing digital forgeries by detecting traces of resampling.
\newblock {\em Trans. Sig. Proc.}, 53(2):758–767, Feb. 2005.

\bibitem{splicecopymove}
Y. {Rao} and J. {Ni}.
\newblock A deep learning approach to detection of splicing and copy-move
  forgeries in images.
\newblock In {\em 2016 IEEE International Workshop on Information Forensics and
  Security (WIFS)}, pages 1--6, 2016.

\bibitem{robustmfjpeg}
Wuyang Shan, YaoHua Yi, JunYing Qiu, and AiGuo Yin.
\newblock Robust median filtering forensics using image deblocking and filtered
  residual fusion.
\newblock {\em IEEE Access}, PP:1--1, 01 2019.

\bibitem{contrast_forgery}
Matthew Stamm and KJ~Ray Liu.
\newblock Blind forensics of contrast enhancement in digital images.
\newblock In {\em 2008 15th IEEE International Conference on Image Processing},
  pages 3112--3115. IEEE, 2008.

\bibitem{inception}
Christian Szegedy, Wei Liu, Yangqing Jia, Pierre Sermanet, Scott Reed, Dragomir
  Anguelov, Dumitru Erhan, Vincent Vanhoucke, and Andrew Rabinovich.
\newblock Going deeper with convolutions, 2014.

\bibitem{mfnet}
Hongshen Tang, Rongrong Ni, Yao Zhao, and Xiaolong Li.
\newblock Median filtering detection of small-size image based on cnn.
\newblock {\em Journal of Visual Communication and Image Representation},
  51:162 -- 168, 2018.

\bibitem{pixeldomainadv}
Benedetta Tondi.
\newblock Pixel-domain adversarial examples against cnn-based manipulation
  detectors.
\newblock {\em Electronics Letters}, 54, 08 2018.

\bibitem{transferadv}
Mauro Barni; Kassem Kallas; Ehsan Nowroozi;~Benedetta Tondi.
\newblock On the transferability of adversarial examples against cnn-based
  image forensics.
\newblock 2020.

\bibitem{cameraid}
Amel Tuama, Fr{\'e}d{\'e}ric Comby, and Marc Chaumont.
\newblock Camera model identification with the use of deep convolutional neural
  networks.
\newblock In {\em 2016 IEEE International workshop on information forensics and
  security (WIFS)}, pages 1--6. IEEE, 2016.

\bibitem{contrastenhance}
Pengpeng Yang, Rongrong Ni, Yao Zhao, Gang Cao, Haorui Wu, and Wei Zhao.
\newblock Robust contrast enhancement forensics using convolutional neural
  networks.
\newblock {\em CoRR}, abs/1803.04749, 2018.

\bibitem{psp}
Hengshuang Zhao, Jianping Shi, Xiaojuan Qi, Xiaogang Wang, and Jiaya Jia.
\newblock Pyramid scene parsing network.
\newblock {\em 2017 IEEE Conference on Computer Vision and Pattern Recognition
  (CVPR)}, Jul 2017.

\end{thebibliography}
}

\end{document}